\documentclass[acmtog]{acmart}

\usepackage{booktabs} 

\usepackage{multirow}
\usepackage{subfig}
\usepackage{wrapfig}

\usepackage{amsmath,amsfonts}
\usepackage{algorithm}
\usepackage{algpseudocode}

\usepackage{array}
\usepackage{textcomp}
\usepackage{stfloats}
\usepackage{url}
\usepackage{verbatim}
\usepackage{graphicx}
\usepackage{hyperref}
\usepackage{soul}
\usepackage{color, xcolor}
\usepackage{tcolorbox}

\setcopyright{acmcopyright}

\definecolor{DeltaColor}{rgb}{0.039,0.73,0.71}
\definecolor{SetaColor}{rgb}{0.867, 0.0235, 0.376}
\definecolor{SigmaColor}{rgb}{0.98,0.45,0.0}
\definecolor{RedColor}{rgb}{0.8,0,0}
\definecolor{AlphaColor}{rgb}{1.0, 0.4, 0.8}
\definecolor{BetaColor}{rgb}{0.8,0,0.8}
\definecolor{GammaColor}{rgb}{0.0,0,0.7}
\definecolor{EpsilonColor}{rgb}{0.353,0.725,0.906}
\definecolor{TauColor}{rgb}{0.423,0.235,0.192}
\definecolor{Black}{rgb}{0,0,0}

\begin{document}

\title{LLM-enhanced Scene Graph Learning for Household Rearrangement}

\author{Wenhao Li}
\authornote{Equal contribution}
\affiliation{%
 \institution{National University of Defense Technology}
 \country{China}}

\author{Zhiyuan Yu}
\authornotemark[1]
\affiliation{%
 \institution{National University of Defense Technology}
 \country{China}}

\author{Qijin She}
\authornotemark[1]
\affiliation{%
 \institution{National University of Defense Technology}
 \country{China}}

\author{Zhinan Yu}
\affiliation{%
 \institution{National University of Defense Technology}
 \country{China}}

\author{Yuqing Lan}
\affiliation{%
 \institution{National University of Defense Technology}
 \country{China}}

\author{Chenyang Zhu}
\affiliation{%
 \institution{National University of Defense Technology}
 \country{China}}

\author{Ruizhen Hu}
\authornote{Corresponding author}
\affiliation{%
 \institution{Shenzhen University}
 \country{China}}

\author{Kai Xu}
\authornotemark[2]
\affiliation{%
 \institution{National University of Defense Technology}
 \country{China}}


\begin{teaserfigure}
  \includegraphics[width=\textwidth]{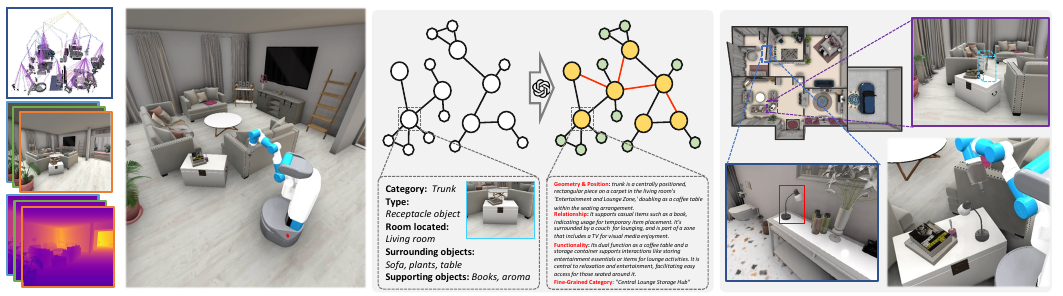}
  \caption{Entering an indoor scene, the robot first performs an RGB-D reconstruction and scene graph extraction. It then transforms the initial scene graph into an affordance-enhanced graph (AEG) and plans actions for finishing the scene rearrangement task, both through prompting a multi-modal foundation model with the scene graph and selected RGB keyframes. With AEG, the robot can detect misplaced objects and determine a proper place for each of them.}
  \Description{bad teaser.}
  \label{fig:teaser}
\end{teaserfigure}

\begin{abstract}
The household rearrangement task involves spotting misplaced objects in a scene and accommodate them with proper places. It depends both on common-sense knowledge on the objective side and human user preference on the subjective side. In achieving such a task, we propose to mine object functionality with user preference alignment directly from the scene itself, without relying on human intervention. To do so, we work with scene graph representation and propose LLM-enhanced scene graph learning which transforms the input scene graph into an affordance-enhanced graph (AEG) with information-enhanced nodes and newly discovered edges (relations). In AEG, the nodes corresponding to the receptacle objects are augmented with context-induced affordance which encodes what kind of carriable objects can be placed on it. New edges are discovered with newly discovered non-local relations. With AEG, we perform task planning for scene rearrangement by detecting misplaced carriables and determining a proper placement for each of them. We test our method by implementing a tiding robot in simulator and perform evaluation on a new benchmark we build. Extensive evaluations demonstrate that our method achieves state-of-the-art performance in misplacement detection and the following rearrangement planning. 
\end{abstract}

%
%


%
%


\maketitle

\section{Introduction}

Scene analysis is a long-standing problem in computer graphics and vision~\cite{patil2024advances} and finds many applications ranging from scene modeling/generation~\cite{fisher2015activity} and VR/AR~\cite{li2021synthesizing} to various robotic applications~\cite{zhang20233d}. Among the applications, some, e.g., robot-operated household rearrangement or room tiding, require a much more in-depth analysis of the scenes than others (e.g., object goal navigation), due to the involvement of both common-sense knowledge on the objective side and human user preference on the subjective side.

When conducting a room tiding task, the agent needs to spot the misplaced objects and accommodate them in proper places~\cite{sarch2022tidee}. This is quite challenging, especially in an unseen environment, as it demands a nuanced understanding of each object's functionality. The determination of object functionality often requires a high-order analysis that considers a broader context~\cite{hu2018functionality}. Moreover, object functionality is often hinged on the house owner's taste or preference, i.e., the owner's intent on how an object is being used. For example, the owner could take a cabinet either as a closet or as a bookshelf, depending on where it is located and mainly what objects are placed in. By analyzing the positioning and context of objects, we can infer the homeowner's preferences and accurately determine the objects' functionality in real-world indoor scenarios.

Large Language Models (LLMs) have demonstrated remarkable effectiveness in challenging zero-shot task planning tasks, including scene rearrangement~\cite{kant2022housekeep,wu2023tidybot} due to their strong common-sense reasoning capabilities. However, LLMs inherently lack specific knowledge about object functionality and placement tailored to the scene at hand. Therefore, scene grounding is crucial to ensure the practicality and relevance of the plans they generate~\cite{rana2023sayplan}.

Another important issue to consider is that LLM ignores human preference totally if the prompt is not carefully designed or tuned.
In the work of TidyBot~\cite{wu2023tidybot}, this is alleviated by explicitly injecting user preference as exemplar placements via prompts and having the LLM summarize user preference out of the exemplars. Note that the exemplar placements need to be specific to the scene in consideration to make the summarization scene grounded, making the exemplar collection tedious. 
\citeN{han2024llm} propose an iterative self-training method to align the LLM planner with user preferences, which again relies on user feedback. On the other hand, we posit that the existing layout of an indoor scene already encapsulates sufficient human preferences, which can be discerned by analyzing the scene's context, providing a natural prompt for the agent to perform rearrangement tasks.

In this work, we propose to mine object functionality with user preference alignment directly from the scene itself, without relying on human intervention. With our method, the user instruction can be as simple as ``tide the house'' and the agent can conduct scene rearrangement automatically based on an in-depth analysis of object functionality and proper reasoning of user preference. One example is shown in Figure~\ref{fig:teaser}. 
Based on our deep analysis of personalized functionality, it appears the user would prefer utilizing the trunk as an unconventional surface for a table. Although this may deviate from the traditional use of a standard trunk, it aligns perfectly with the user's unique desire for a more tailored and comfortable space.

To achieve that, we propose LLM-enhanced in-depth scene analysis to extract affordance for each carriable object in the scene reflecting simultaneously common-sense knowledge and personalized preference, both with scene grounding. To do so, we work with scene graphs which is an LLM-friendly representation for scene analysis. Starting with a vanilla scene graph, we harness LLM to enhance it into an affordance-enhanced graph (AEG) with information-enhanced nodes and newly discovered edges (relations). In particular, we construct a prompt containing the initial scene graph (in XML format) and the RGB-D observations for a multi-modal foundation model which we use GPT4V in this work. GPT4V then gradually converts the scene graph into an AEG where the nodes of receptacle objects are augmented with context-induced affordance mined by GPT4V. In particular, context-induced affordance encodes what kind of carriables can be placed on it. Some new edges are discovered based on analyzed non-local relations.

With AEG, we perform scene rearrangement in two steps. The first step is misplacement detection. For each carriable, we feed its context-induced affordance, its associated receptacle objects, and the task instruction into GPT4V. GPT4V then outputs for each carriable a probability of misplacement. In the second step, we choose the top few most probably misplaced carriables and query GPT4V for potential receptacles for correct placement. A straightforward method for receptacle query is to put all candidate receptacles in the scene into a very long prompt which may cause degraded accuracy and even hallucinations in output. We, instead, opt to pre-compile task-related affordance (affordance with task description) of all receptacles into an external memory in a similar spirit to Retrieval Augmented Generation (RAG). This way, the prompt for receptacle query is composed of a few most relevant entries retrieved from the memory, thus significantly improving query quality.

\begin{figure*}[t]
\centering{\includegraphics[width=\textwidth]{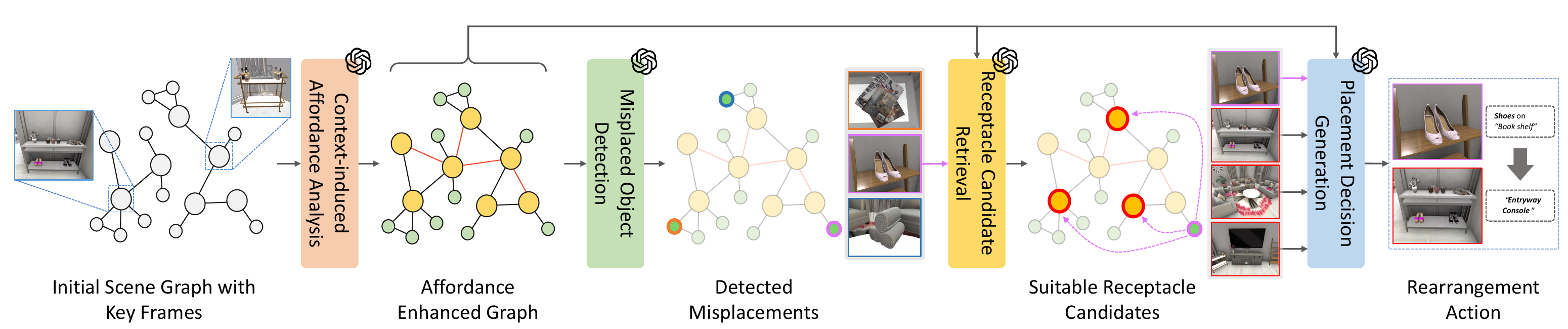}}
\caption{Given a scene graph (SG) with key frames, we utilize Large Language Model (LLM) to perform context-induced affordance analysis for all objects in the scene. These affordances are incorporated into the SG, updating the nodes and edges to construct an Affordance Enhanced Graph (AEG).
We then evaluate the appropriateness of the current placements of all carriable objects in the AEG based on the affordance information of the objects and their receptacles, identifying misplaced items. For each misplaced carriable, we rate the suitability of each receptacle in the AEG as a placement target. The top k suitable receptacles are selected as candidates, and their affordances are retrieved as prompts to the LLM to generate the placement decision.}
\label{fig:pipeline}
    
\end{figure*}


We test our method by implementing a robot system for indoor scene rearrangement in the Habitat 3.0 simulator~\cite{puig2023habitat}. Entering an indoor scene, the robot first performs an RGB-D reconstruction and extracts a vanilla scene graph for the scene. It then transforms the scene graph into an AEG and plans actions for the rearrangement task automatically by prompting GPT4V with the scene graph and selected RGB keyframes. 
We also contribute a benchmark for evaluating the performance of scene rearrangement through annotating the Habitat Synthetic Scenes Dataset (HSSD 200)~\cite{khanna2023hssd} with ground-truth receptacles for each carriable. Extensive evaluations of these two benchmarks demonstrate that our method achieves state-of-the-art performance on misplacement detection and the following rearrangement planning in the household rearrangement task.

Our main contributions include:
\begin{itemize}\vspace{-6pt}
    \item We introduce LLM-enhanced scene graph learning for deep mining of object functionality with user preference alignment, which results in an affordance-enhanced graph (AEG) encoding context-induced affordances of receptacles. 
    \item Based on AEG, we propose LLM-assisted misplacement detection and correct placement determination, with carefully designed mechanisms for improved efficiency.
    \item We provide a benchmark of scene rearrangement by annotating a scene dataset with ground-truth receptacles. With the benchmark, we test our method with an end-to-end robot system for scene rearrangement implemented in simulator.
\end{itemize}

\section{Related Works}

\paragraph{Scene understanding and affordance analysis}
Scene understanding is an important area of research that can be broadly categorized into perception tasks ~\cite{georgakis2017synthesizing, silberman2012indoor, zhang2020fusion} and application tasks ~\cite{achlioptas2020referit3d, azuma2022scanqa, duan2022disarm}. As the field of embodied AI develops, scene understanding research has become more focused on understanding how scenes support the behavior of embodied agents such as human avatars and robots, i.e. affordance ~\cite{gibson1977theory}. 

Affordance prediction has been approached at both the object level and the scene level. Object-level methods focus on understanding how humans use items ~\cite{nagarajan2019grounded,koppula2014physically,fang2018demo2vec, deng20213d, nagarajan2020ego}, while scene-level research emphasizes how to naturally place a human avatar within a scene ~\cite{wang2024move, li2019putting, gupta20113d, kulal2023putting, grabner2011makes, zhu2016inferring, savva2014scenegrok, savva2016pigraphs}. Different from the aforementioned works which analyze affordance from a human interaction aspect, some studies focus on the analysis of object-object affordance ~\cite{mo2022o2o, tang2021relationship, li2022ifrexplore}. Their methods predict the physical interaction between objects, such as placement and stacking. ~\cite{tang2021relationship, li2022ifrexplore} extract potential object interactions within a given scene. However, all these object-object affordance works do not consider the ''scene affordance'' of the objects, which requires an in-depth analysis of factors including the object's geometry, pose, and the scene context, such as its position in the scene and the objects already interacting with it. 



\paragraph{Foundation models for scene understanding}
Visual language models ~\cite{radford2021learning} and large language models ~\cite{achiam2023gpt} have been recently used in scene understanding tasks ~\cite{conceptfusion, hong20233d, wang2023chat}. Some works~\cite{li2023one, nguyen2023open, van2023open, zhang2024rail} use or distill the rich commonsense provided by foundation models to identify affordance for a sole object.
~\cite{rana2023sayplan, rajvanshi2023saynav, ni2023grid} use the scene graph to ground LLMs for navigation task. However, they just use the scene graph as a map without extracting object interaction information hiding in the scene graph. While in our work, we focus on using scene graph to ground LLMs in analyzing the object affordances, the analysis results are then sent back to the scene graph for deeper analysis.

\paragraph{Rearrangement}
Rearrangement is a special embodied AI task to test the robots' intelligence ~\cite{batra2020rearrangement}.
In a given initial state of the scene, robots are tasked with manipulating the poses of objects to align with desired goal states. These goal states can be represented through various means such as object pose transformations~\cite{liu2021ocrtoc}, language instructions~\cite{liu2022structformer, structdiffusion2023}, or visual images ~\cite{weihs2021visual, huang2023efficient}. 
It is worth noting that in certain cases like household tidying, robots may need to rearrange objects based on the original scene and their experiences. This necessitates the robot to have human-like common sense and the ability to decipher user preferences embedded within the original scenarios. Previous works in household rearrangement rearrange objects without considering the fine-grain affordances constrained by the scene ~\cite{kant2022housekeep, sarch2022tidee}. In our method, we analyze the object affordance with scene graph and LLMs, which enable us rearrange the households while follow the original functionality of the object as much as possible.
\section{Method}

Figure~\ref{fig:pipeline} presents an overview of our approach. Starting with a 3D scene, abstracted into a scene graph where each object node is linked to a key frame, our method initiates by harnessing the contextual details of each node. This information is used to augment the graph, infusing it with induced affordances and uncovering new semantic connections between objects.
Adopting the objective of "tidying the house," the affordance-enriched graph serves as the foundation for identifying objects that are out of place. To determine the most suitable location for each misplaced object, our system initially extracts a set of potential receptacle candidates from the graph. These candidates are then meticulously evaluated and compared to arrive at the optimal placement decision.
It is important to highlight that all the pivotal steps of our methodology are executed in a zero-shot learning context, leveraging the advanced capabilities of the Large Language Model (LLM), specifically GPT4V. 

\begin{figure*}[t]
\centering{\includegraphics[width=0.98\textwidth]{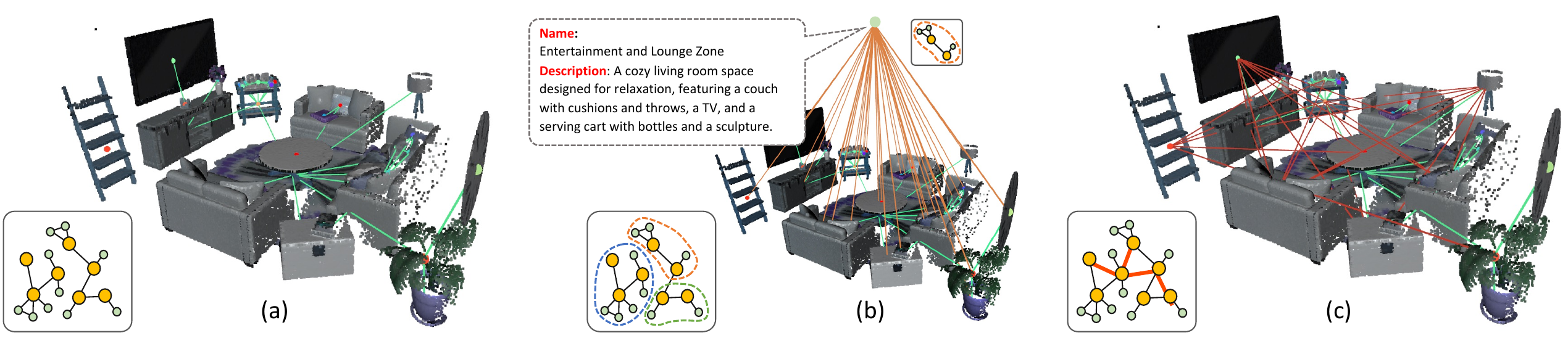}}
\caption{
Context-induced affordance analysis for objects in the scene graph. (a) For each object, we conduct a local analysis based on its textual contents related to its neighborhood in the graph and its key frame image, storing the results in the nodes as local affordances. (b) We construct object-area-room hierarchies and aggregate the local affordances of all objects within an area. Using LLM, we summarize the content and functionality of the entire area based on the aggregated information, obtaining descriptions for all the areas. (c) We use the descriptions of all areas in a room as the global context and assign them to all nodes within the room to allow the LLM to identify meaningful semantic context information for each object. With this meaningful context, we construct new semantic edges between objects in the same room and update the affordances, completing the context-induced affordance analysis of the scene graph.}
\label{fig:global}
\end{figure*}

 
\begin{figure}[t]
\centering{\includegraphics[width=0.5\textwidth]{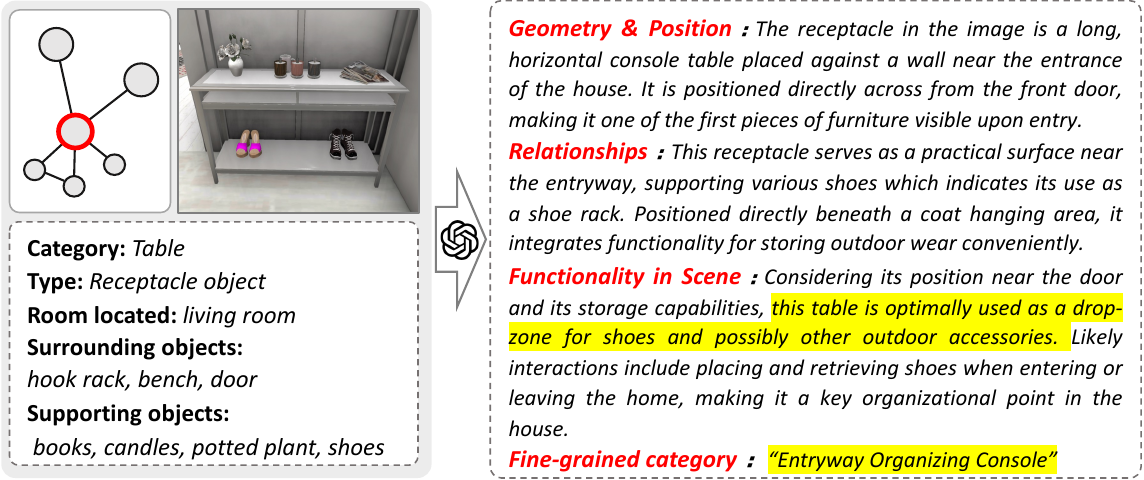}}
\caption{Local affordance analysis of an example receptacle. LLM analyzes the contextual details from the scene graph and visual image to determine the potential functionality of the receptacle, assigning a more specific category.
}
\label{fig:local}   
\end{figure}

\subsection{Context-induced Affordance Analysis}



In this section, we will explain how the initial scene graph is enhanced by using LLM to analyze the contextual information to in a local-to-global manner, as shown in Figure~\ref{fig:global}.

\paragraph{Initial scene graph}
To simplify the problem setting, we assume that a vanilla scene graph $S$ has already been constructed from an RGBD scanning sequence for the given scene, where each node represents an object instance in the scene and two nodes will be connected with an edge if they are close enough. 
Moreover, several attributes of the object instance are stored in each node, including the \emph{instance ID}, \emph{category label}, \emph{located room label}, and \emph{rearrangement type (carriable, receptacle, or others)}, where the \emph{rearrangement type} follows the definitions from the Vocabulary Mobile Manipulation (OVMM)~\cite{homerobotovmm} Challenge. 
In the task, only carriable objects can be rearranged,  receptacle objects serve as targets for placing carriable objects, and other objects do not interact with the agent during the task but can provide contextual information for other objects.
The relationship between two nearby objects is also categorized into three different types, including \emph{near}, \emph{on}, and \emph{support}, and stored on the connecting edge. Note that if a scene graph is not given, it can be constructed using established methods, with detailed guidance available in the supplementary material.


To help with the inference, we further associate each node with a representative RGB frame selected from the input RGBD scanning sequence to provide a visual context.   
In more detail, for each object instance, we first calculate the number of its projected pixels across all RGB-D images, creating an object-to-image-pixel matrix ($N*M$), where $N$ is the number of the object instance and $M$ is the number of frames.
To select the most informative frame, we identify all the neighboring objects and sum up their projected pixel numbers together with that of the current instance on each frame.
The frame with the maximal number of projected pixels is selected as the key frame for the node, as it contains the most visual information of the surrounding context of the instance to be considered.  

\paragraph{Local context analysis}

Local context refers to the neighboring information stored in the scene graph $S$. 
For each receptacle and other object in the $S$, we input all the textual contents related to its neighborhood and structure it using a predefined template. This template, along with its associated keyframe, forms the basis of a prompt designed for a Large Language Model (LLM) to analyze the object's functionality through Chain of Thought reasoning~\cite{wei2022chain}. Consequently, the output is systematically formatted to detail the object in its \emph{Geometry \& Position, Relationships, Functionality,} and \emph{Fine-grained Category}.


Figure~\ref{fig:local} shows the local affordance analysis result of an example receptacle.  
In this example, the LLM first analyzes the geometric features of a table based on the visual information provided by the corresponding image. 
It combines this information with contextual information given by the scene graph to infer that the table is positioned against the wall at the entrance of the living room. Then, leveraging the position information and the supporting objects of the table in the scene graph, the LLM infers that the table can be likely used for placing outdoor wear.
In the functionality module, based on the preceding analysis, the LLM further deduces that the table is suitable for placing items such as shoes that are typically found near entrances, as well as keys and other items that are temporarily stored when entering or leaving. 
As a conclusion, the LLM assigns a new fine-grained category, "Entryway organizing console", to the table based on the above analysis.

\paragraph{Global context analysis}
Although our vanilla scene graph $S$ can effectively describe the surrounding context of a receptacle, 
the functionality of the receptacle may also depend on objects that are far away thus without connecting by an edge in $S$.
For example, a sofa and a TV have a strong functional relationship (people sit on the sofa to watch TV), but they are typically arranged at a considerable distance from each other.
Therefore, to consider more global context in the scene, we first build a hierarchical structure of $S$ and use it to bridge objects that are spatially distant yet functionally interrelated. 
In this way, new semantic edges are added, and the information stored in each object node is also updated, resulting in a new affordance-enhanced graph (AEG), as shown in Figure~\ref{fig:global}.
The details of the hierarchy construction process can be found in the supplementary material. 
Note that objects are first grouped into an intermediate granularity level called ``area'', and then further merged into a room. 
Using this "object-area-room" hierarchy to aggregate contextual information is more efficient than aggregating directly from objects to rooms. This method allows for layered, structured storage of aggregated information, reserving important details.

With the ``object-area-room'' hierarchy, we first use LLM to aggregate contextual information for each area separately. For each area, we collect the node information stored in $S_l$ of all the objects inside the area and design a prompt for LLM to generate the following content based on a fixed template: \emph{Description} and \emph{Fine-grained name}.
Therefore, each area will be associated with the above information together with the list of the objects inside this area. We then integrate all the information from the areas within a room and use it as the global context for the room to extract distant context for all the receptacles and fixed objects within the room. 




For each receptacle, we input the basic spatial information template description, its local affordance analysis as well as the global context of the room to the LLM. With these inputs and a designed prompt, LLM can extract the distant but meaningful context from the global information and add these contexts as semantic edges to the SG. These context outputs will be in a format of \emph{a list of objects, indicating objects within the room that are functionally related to the given receptacle}, as well as \emph{the specific functional relationship between these objects and the given receptacle.}
After obtaining this meaningful context, we add semantic directed edges from the given receptacle to related objects, reflecting their functional relations.


\begin{figure}[t]
\centering{\includegraphics[width=0.5\textwidth]{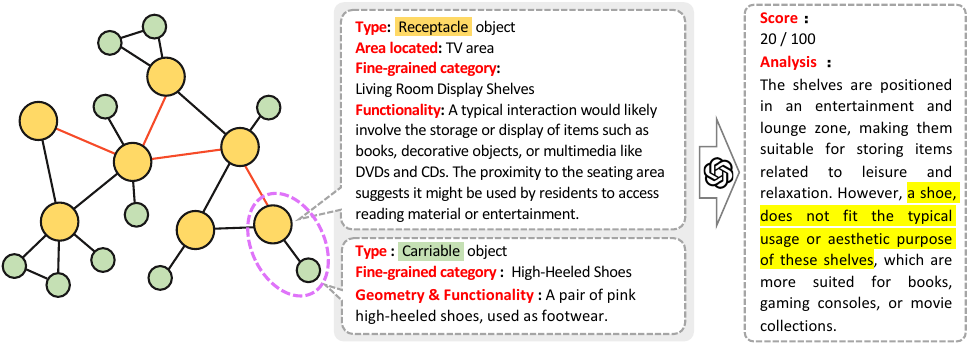}}
\caption{
An example of placement scoring with the LLM. 
We set a fixed standard in the prompt for the LLM to refer to when rating: 100 for perfect placement, 0 for wrong placement, and 50  for placements difficult to judge.
}  
\label{fig:misplace}
\end{figure}

\paragraph{Affordance updation.}
In the end, we use LLM to update context-induced affordance based on the previous analysis, the additional global context, as well as the semantic edges, with the same output format as in the local context affordance. 
In this way, our context-induced 
affordance includes both surrounding spatial context and distant semantic context, resulting in a more comprehensive understanding of receptacle's functionality in this room. 
We refer to the final scene graph with updated node information and newly added semantic edges as Affordance Enhanced Graph (AEG) and use it for further misplacemnet detection as well as rearrangement planning.


\begin{figure*}[t]
\centering{\includegraphics[width=0.98\textwidth]{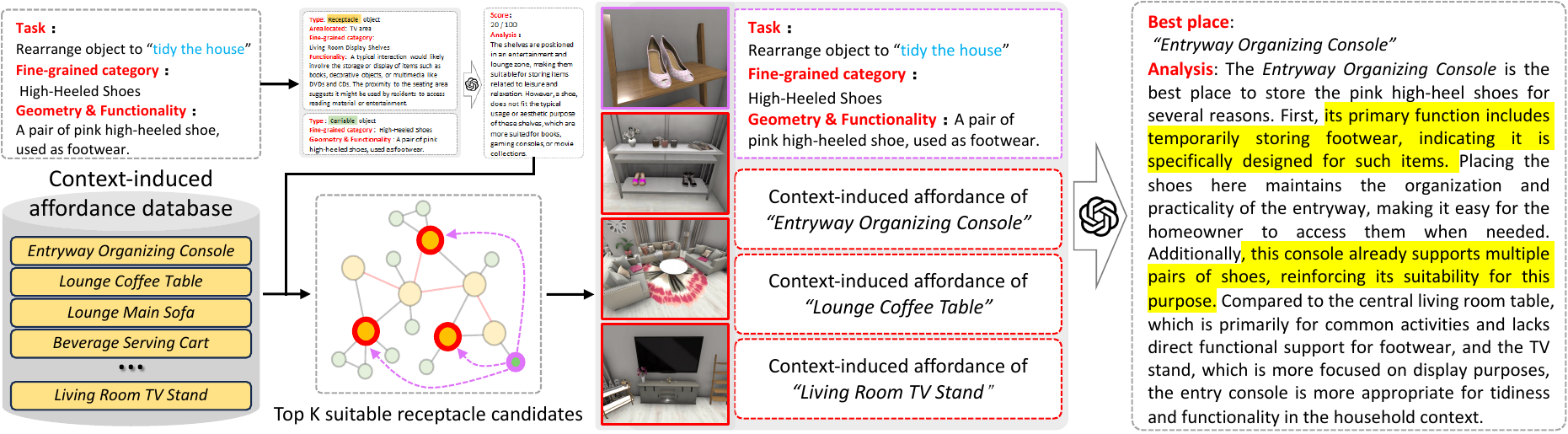}}
\caption{We use a score-based retrieval-augmented method for placement decision generation. First, we construct a context-induced affordances database for all receptacles in the AEG. For each carriable object to be rearranged, we treat the object as a query and evaluate the suitability of each receptacle in the database as a placement target using the LLM scorer. The top k receptacles with the highest scores are selected as candidate receptacles. We retrieve their context-induced affordances from the database and use them along with the query as prompt input to the LLM to generate the best placement plan.}
\label{fig:RAG}
\end{figure*}


\subsection{Misplaced Object Detection}






Since our task is to ``tide the house'', we are looking for carriables that currently have an unreasonable placement and need to be rearranged to complete the task.
To achieve this, we introduced an LLM-based placement scorer to rate the suitability of a placement based on the affordance and description of carriable-receptacle pair. With this placement scorer, we can determine if a carriable object is appropriately placed currently and find all the misplaced carriables. 

Specifically, 
Due to the token limitation of the LLM, we cannot input all the carriable objects at once for scoring, so we score carriables one by one.
For each carriable object, we collect the description of the carriable object, the context-induced affordance of the receptacle where it is currently located, and our task instruction, which is "tide the house." 
With the collected information above as well as a well-designed prompt, LLM will analyze the current placement of the given carriable object to determine the suitability of the placement and assign a $0\textasciitilde100$ score to this relevance.
Figure~\ref{fig:misplace} shows an example of the scoring process. We set 50 points as the threshold and all those carriable objects with scores no more than 50 are considered to be misplaced and should be rearranged. 
These carriables are then reorganized in accordance with their respective scores, arranged in ascending order, to facilitate the planning of subsequent rearrangements.


\subsection{Object Rearrangement Planning}



For each misplaced object, our goal is to find the most suitable receptacle in the scene as the placement target with the context-induced affordance information stored in the AEG. 
To allow LLM to find the most suitable placement, it is necessary to consider all the relevant receptacles for comparison. However, directly inputting the affordances of all possible receptacles in the AEG to LLM will not only flood it with excessive irrelevant information, impairing its reasoning ability but also make it lose focus, increasing the likelihood of hallucinations.
Inspired by Retrieval Augmented Generation (RAG)~\cite{gao2023retrieval}, which retrieves the task-related text chunks from a given database to augment LLM's task-solving ability and prevent hallucination, we construct a database of receptacle affordance and retrieve the top-k task-related receptacles from this database as suitable candidates and send them to LLM to select the best target, as shown in Figure~\ref{fig:RAG}. In this way, the task-irrelevant receptacles 
will be filtered during the retrieving process, and with the rest top-k candidates as prompt, LLM can make accurate placement decisions and generate high-performance rearrangement planning.

\paragraph{Receptacle candidate retrieval.}

Different from the classic RAG, which splits a long text into multiple chunks to create the database and retrieves the most relevant chunks based on semantic similarity to the query, our dataset consists of descriptions of all receptacle nodes in the AEG and the top-k most relevant receptacles are retrieved based on task relevance to the query.
To achieve this, we adopt our LLM scorer from the misplacement detector with a minor change of prompt and use it to rate the appropriateness of placement based on task instruction.
With all placements of receptacle rated, we can choose the top k receptacles with the highest scores as candidates and use their context-induced affordance as retrievals for the placement decision generation.

\paragraph{Placement decision generation.}
We introduce an LLM-based placement decision generator to select the best receptacle among the proposed candidates and generate a rearrangement plan. 
With the top-k rated receptacles, we feed their context-induced affordance as well as the query (description of carriable to rearrange together with the task) into the placement decision generator. 
The LLM then selects the best placement target, outputting the placement decision with a paragraph of analysis.
With the generated placement target, we can pick up all the misplaced carriable objects and put them in their target locations, completing the rearrangement process.

\section{Experiment and Results}

\subsection{Benchmark}

\paragraph{Datasets.}
 Two benchmarks were used for testing, one proposed by Tidybot~\cite{wu2023tidybot} and another dataset annotated by us based on the Habitat Synthetic Scenes Dataset (HSSD 200)~\cite{khanna2023hssd}. 
The TidyBot [Wu et al. 2023] benchmark is centered around learning individual preferences from human-provided examples, which results in smaller scenes with each object having a single ground truth (GT) placement. In contrast, our focus is on inferring commonsense object arrangements in large, house-level scenes, which offer richer contextual information and more varied placement options. To fully assess performance in scenarios that better resemble real-world conditions, we developed a context-driven benchmark comprising 8 scenes from Habitat Synthetic Scenes Dataset (HSSD 200) [Khanna et al. 2023], including 36 rooms, 105 areas, and 188 receptacle objects. We randomly placed carriable objects in these scenes, creating over 4000 messy scenes to evaluate misplacement detection. 
To determine correct object placement, we enlisted skilled individuals to control a robot, navigating each scene to identify the most appropriate receptacle for each new object, with a minimum of one and a maximum of five options. As all scene information was readily accessible, the provided annotations accurately reflect human commonsense about object placement. After gathering all annotations, we calculated the majority vote for each receptacle per object in each scene, generating a ranked list of receptacle choices.


\begin{table}[t]
\caption{Comparison to Tidybot on its benchmark by sorting criteria. We calculated the average success rate of different sorting criteria by weighting the success rate with the number of objects within that criteria.}
\label{tab:comp_tidybot}
\scalebox{0.8}{
\begin{tabular}{@{}l|cccccc@{}}
\toprule
Method  & Category        & Attribute       & Function        & Subcategory     & Multiple        & Average         \\ \midrule
Tidybot & 91.0\%          & 85.6\%          & \textbf{93.9\%} & \textbf{90.1\%} & 93.5\%          & 90.7\%          \\
Ours    & \textbf{91.3\%} & \textbf{88.8\%} & 93.7\%          & 89.1\%          & \textbf{96.0\%} & \textbf{91.4\%} \\ \bottomrule
\end{tabular}
}
\vspace{-10pt}
\end{table}

\begin{table*}[t]
\caption{Comparison with Housekeep \cite{kant2022housekeep} on Rearrangement and Misplacement detection task. The random selection strategy is added in the comparison as a baseline to demonstrate improvement given by the different methods. The best results for each indicator are highlighted in bold.}
\label{tab:comparison}
\resizebox{0.9\linewidth}{!}{
\begin{tabular}{@{}l|cccccccc|cccc@{}}
\toprule
\multirow{2}{*}{Method} & \multicolumn{8}{c|}{Rearrangement (Top k NDCG)}                                                                                       & \multicolumn{4}{c}{Misplacement detection}                                                                               \\
                        & 1              & 2              & 3              & 4              & 5              & 6              & 7              & 8              & \multicolumn{1}{c}{Accuracy} & \multicolumn{1}{c}{Recall} & \multicolumn{1}{c}{Precision} & \multicolumn{1}{c}{F1 score} \\ \midrule
Random                  & 0.181          & 0.193          & 0.204          & 0.216          & 0.234          & 0.253          & 0.274          & 0.295          & 0.57                         & 0.594                      & 0.815                         & 0.687                        \\
HouseKeep~\cite{kant2022housekeep}              & 0.265          & 0.249          & 0.268          & 0.286          & 0.302          & 0.316          & 0.333          & 0.351          & 0.637                        & 0.707                      & 0.831                         & 0.764                        \\
Ours                    & \textbf{0.648} & \textbf{0.627} & \textbf{0.621} & \textbf{0.635} & \textbf{0.652} & \textbf{0.667} & \textbf{0.684} & \textbf{0.696} & \textbf{0.829}               & \textbf{0.88}              & \textbf{0.908}                & \textbf{0.894}               \\ \bottomrule
\end{tabular}
}
\end{table*}

\paragraph{Metrics.}
For the household rearrangement task, we evaluated our method mainly in two aspects: misplacement detection and carriable object rearrangement planning.
When the ground-truth object placements are available, we measure the detection performance based on accuracy, recall, precision, and F1 score following most existing works~\cite{padilla2020survey}. 
In our benchmark tests, each object may have multiple placement positions, with a ranking that reflects human common sense. To more comprehensively compare the performance of different methods in carriable object rearrangement planning, we also require these methods to output a ranking of recommended placement positions. Normalized Documented Cumulative Gain, which is regarded as a measure of how close a ranking is to the ground truth ranking, can assess whether the methods can select appropriate placement positions based on human common sense in the rearrangement planning. In practice, we take the best 8 placements for each carriable object to calculate the NDCG.

\subsection{Quantitative Comparisons}

\paragraph{Comparison on Tidybot benchmark.} We first compared our method with the Tidybot~\cite{wu2023tidybot} on its benchmark for the rearrangement planning task. 
Note that the Tidybot needs to imitate the given preferences to perform the planning. Two demonstration placements are provided for each receptacle for imitation in the Tidybot's test. 
On the other hand, we only take the given demonstration placements as a regular context for affordance analysis during testing. 
As shown in Table~\ref{tab:comp_tidybot}, our method can still achieve better performance than Tidybot under this unfair setting, which demonstrates that our AEG can support LLMs in extracting effective information even when the context information is relatively sparse.

\paragraph{Comparison on our context-oriented benchmark.} 
Note that Tidybot fails when testing on our context-oriented benchmark due to its limited performance in large-scale scenes and the setting where no certain demonstration placements are provided. Thus, we compare our method with alternative Housekeep~\cite{kant2022housekeep}, 
a state-of-the-art learning-based rearranging method that extracts priors from the joint probability distribution among rooms, receptacles, and objects provided by their human-annotated dataset. Unlike our method, the approach does not consider the functionality implied by the spatial relationships of surrounding objects and the objects' geometries.
As shown in the comparison results in Table~\ref{tab:comparison}, our approach significantly outperforms Housekeep in both placement rearrangement and misplacement detection tasks. 
It demonstrates that the proposed AEG can significantly better enhance the agent's understanding of scenes by leveraging additional context such as spatial, geometric, and visual cues besides semantic relationships.

\subsection{Ablation Studies}

\begin{table}[t]
\caption{Ablation study on different context-induced types via evaluating the misplacement detection performance on our context-oriented dataset. 
}
\label{tab:misplace_ablation}
\resizebox{0.9\linewidth}{!}{
    \begin{tabular}{@{}ccc|cccc@{}}
    \toprule
    \multicolumn{3}{c|}{Context-induced types} & \multicolumn{1}{c}{\multirow{2}{*}{accuracy}} & \multicolumn{1}{c}{\multirow{2}{*}{recall}} & \multicolumn{1}{c}{\multirow{2}{*}{precision}} & \multicolumn{1}{c}{\multirow{2}{*}{F1}} \\
    Visual       & Textual       & Global       & \multicolumn{1}{c}{}                          & \multicolumn{1}{c}{}                        & \multicolumn{1}{c}{}                           & \multicolumn{1}{c}{}                    \\ \midrule
                 &               &              & 0.774                                         & 0.813                                       & 0.901                                          & 0.852                                   \\
    \checkmark            &               &              & 0.811                                         & 0.856                                       & 0.907                                          & 0.880                                   \\
                 & \checkmark             &              & 0.821                                         & 0.878                                       & 0.899                                          & 0.888                                   \\
                 \checkmark            & \checkmark             &              & 0.828                                         & 0.878                                       & \textbf{0.909}                                          & 0.892                                   \\
                 \checkmark            & \checkmark             & \checkmark            & \textbf{0.829}                                & \textbf{0.880}                              & 0.908                                 & \textbf{0.893}                          \\ \bottomrule
    \end{tabular}
    }
    \end{table}

\begin{figure}[t]
\centering
\includegraphics[width=0.98\linewidth]{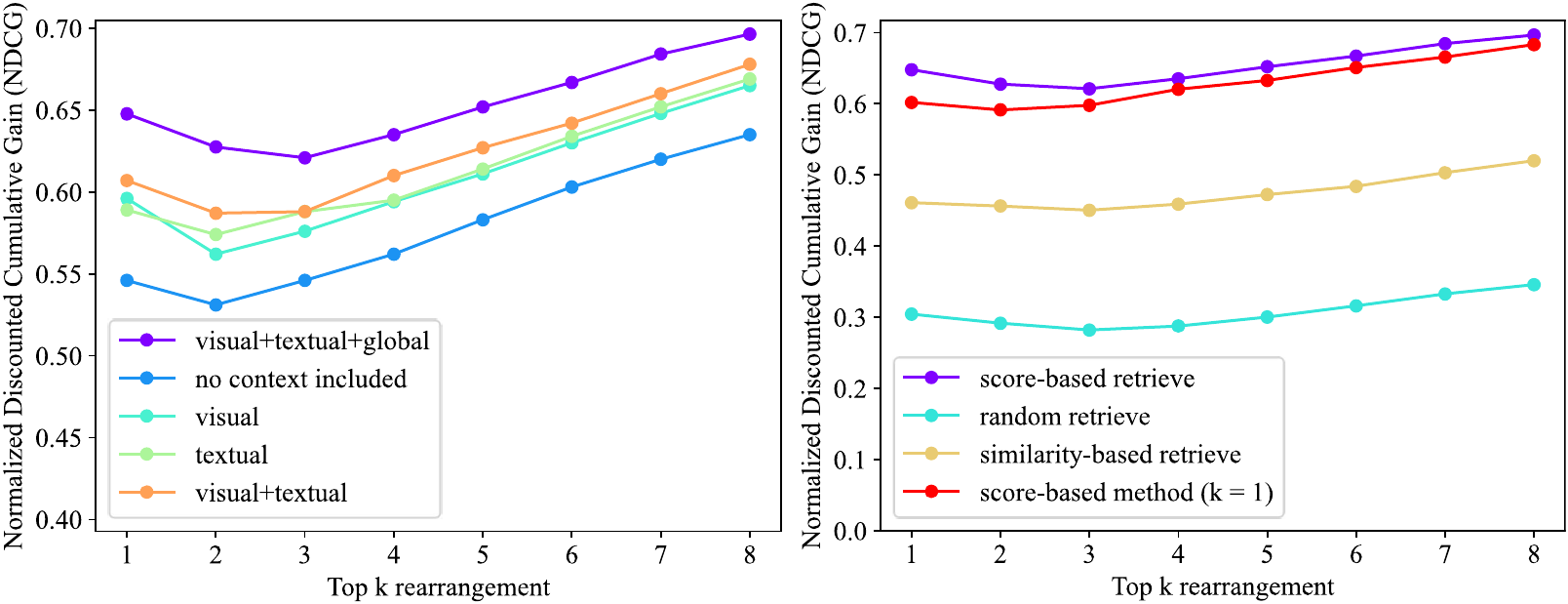}
\vspace{-10pt}
\caption{Ablation study on context-induced affordance enhancement and LLM-based scorer via measuring the NDCG performance. 
}
\label{fig:ablation_ndcg}
\vspace{-15pt}
\end{figure}

\paragraph{Necessity of context-induced affordance.} The key to the success of our method lies in the affordance-enhanced graph (AEG) derived via context-induced affordance analysis. 
To verify the necessity of AEG with full context information, we compare it with other alternatives using \emph{a vanilla scene graph} (no local context analysis), \emph{a visual-context-enhanced scene graph} (only keyframe context is given), or \emph{a textual-context-enhanced scene graph} (only textual description is given) on our benchmark.
The results are presented in Table~\ref{tab:misplace_ablation} and Figure~\ref{fig:ablation_ndcg} (left). 
We can see that the vanilla scene graph can not provide enough support for both tasks since it can not reflect common-sense knowledge and personalized preference in the scene. Meanwhile, we find that the visual context is more effective for the misplacement detection and the textual context is more essential for the rearrangement. Our method takes both into the affordance analysis and achieves the best performance in both tasks.


\paragraph{Necessity of LLM score-based retriever.} Compared to asking the LLM to predict the final receptacle directly based on AEG, we find that adopting an LLM scorer to retrieve candidates for a RAG query can provide more reasonable results. 
To make a fair evaluation, we compare our LLM score-based retrieval strategy with a similarity-based candidate retrieval method~\cite{ragsurvey} that is widely used in NLP applications. 
We also added a random selection strategy that serves as a baseline. 
As presented in the right of Figure~\ref{fig:ablation_ndcg}, our LLM score-based retriever can provide significantly better performance than other alternatives. We can effectively filter out a large number of irrelevant options with the LLM-based scoring, which greatly reduces the difficulty for the LLM in making the decision.

We also implemented ablations about key frame selection, number of retrievals, and score prompting in the supplementary material.

\subsection{Qualitative Results}
\begin{figure*}[t]
  \includegraphics[width=0.85\textwidth]{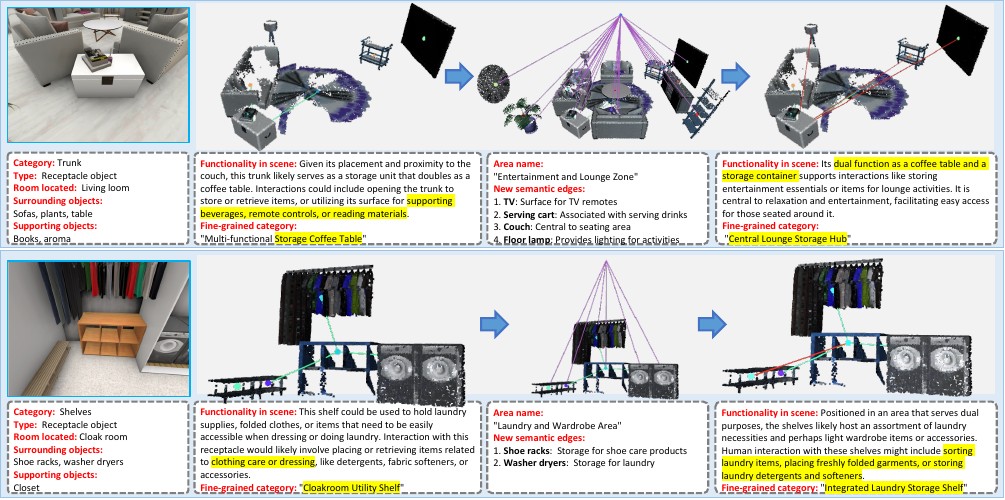}
  \caption{Context-induced graph enhancement examples. Top: Through analyzing the local context, a fine-grained category "multi-functional storage coffee table" is predicted for the trunk. While a global context such as the area name "lounge zone" is given, the affordance of this trunk can be further refined as a "central lounge storage hub". Bottom: The local context analysis of the given shelf infers that this shelf is a "cloakroom utility shelf" based on the surrounding clothes. The area information adopted in the next global context analysis can revise this fine-grained label as "integrated laundry storage shelf" more precisely.} 
  \Description{result v2}
  \label{fig:v2-1}
\end{figure*}

\begin{figure*}[t]
  \includegraphics[width=0.85\textwidth]{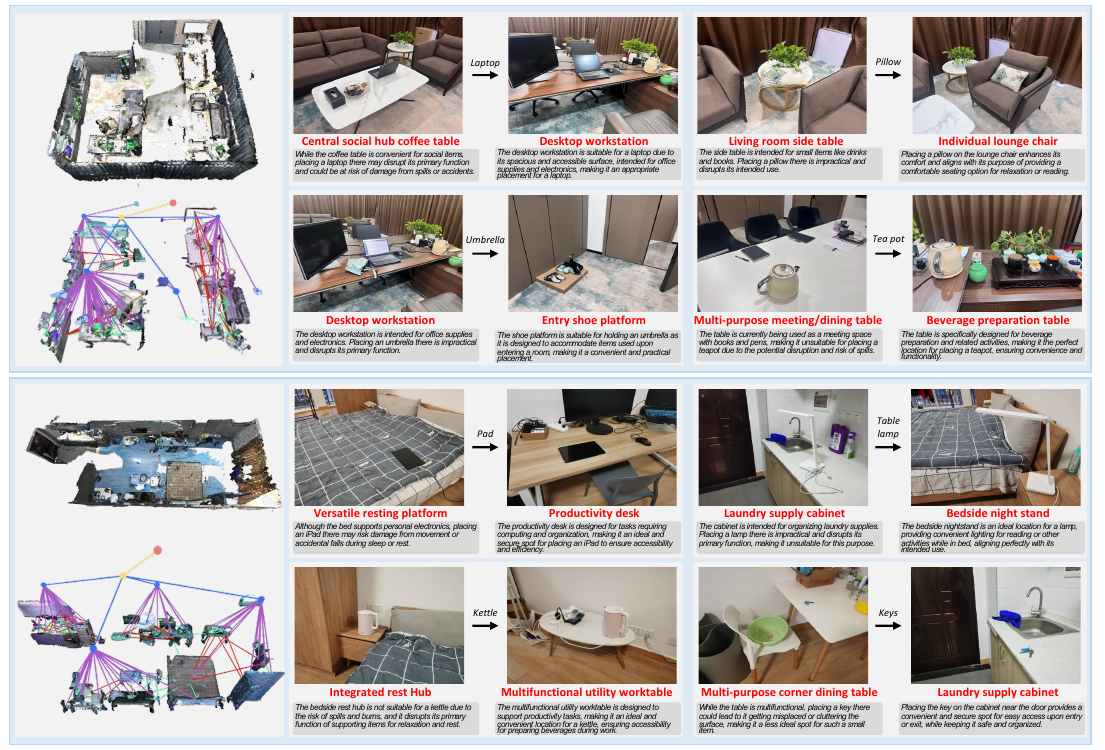}
  \caption{Rearrangement planning examples in the real scenes, demonstrating eight rearrangement cases in two scenes. For example, the desk in the top scene is used as a workstation due to the placement of a monitor and laptop on top of it. Therefore, the umbrella on it is detected as a misplaced object and our method finds a more proper receptacle ``Entry shoe Platform'' for placement.}
  \Description{result v2}
  \label{fig:v2-2}
\end{figure*}

\begin{figure*}[t]
  \includegraphics[width=0.96\textwidth]{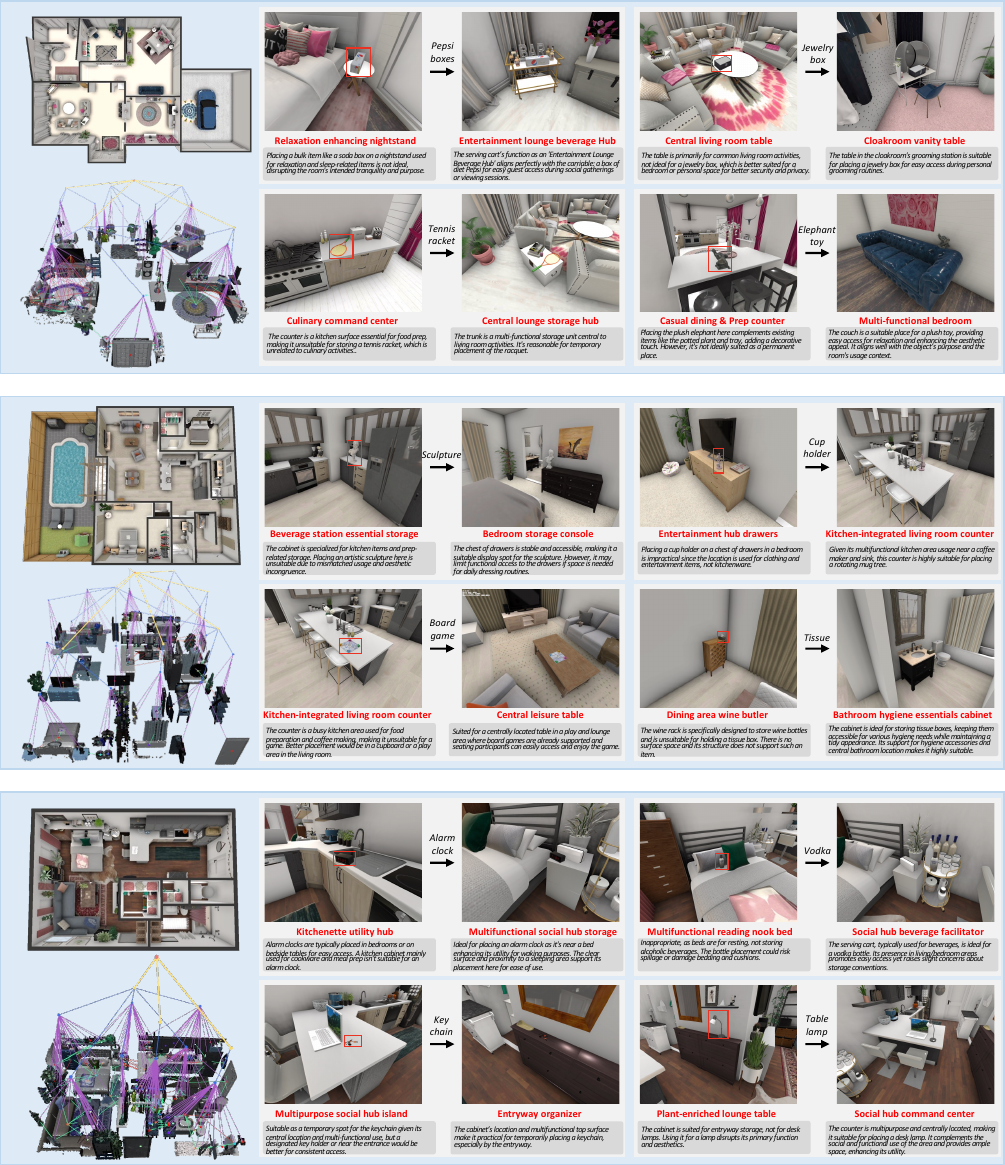}
  \caption{Rearrangement planning examples in the synthetic scenes, demonstrating twelve rearrangement cases in three scenes. Here is an interesting example in the bottom scene, where our agent finds a better placement, the entryway organizer, for the key chain compared to its original location. The insight here is that our context analysis indicates that the desk is only suitable as a temporary spot for the keychain given its central location and multi-functional use. Therefore, our method searches the entry area of the scene and finds a more proper receptacle for placing the keychain.}
  \Description{result v1}
  \label{fig:v1}
\end{figure*}

To better illustrate the pipeline of our method, we show some examples of context-induced affordance enhancement in Figure~\ref{fig:v2-1}. 
Note how the agent derives more accurate functional descriptions of target objects than the original general semantic tags.
Besides, we showcase several examples of carriable object placement planning based on AEG in Figure~\ref{fig:v2-2} with real-world scenarios and Figure~\ref{fig:v1} with synthetic environments.
We can see that the agent can plan object placements that better fit the scene context based on AEG.

\section{Conclusion}

We introduced a method to automatically mine object functionalities aligned with user preferences from a scene by using multi-modal foundation models prompted with scene graphs and RGB keyframes. The resulting affordance-enhanced graph effectively detects misplaced objects and determines their proper locations.
Our affordance enhancement, which integrates contextual information, is contingent upon the accuracy of the initial scene graph, thus it cannot rectify errors inherent in its construction. Moving forward, we are keen on pursuing the joint optimization of the initial scene graph through end-to-end and/or active learning approaches.

\clearpage
\section*{Acknowledgment}
This work was supported in part by the NSFC (62325211, 62132021, 62322207, 62372457), the Major Program of Xiangjiang Laboratory (23XJ01009), the Natural Science Foundation of Hunan Province of China (2021RC3071) and Shenzhen Science and Technology Program (RCYX20210609103121030).

\bibliographystyle{ACM-Reference-Format}
\bibliography{main.bib}

\clearpage


\clearpage
\appendix
\section{Rearrangement Setting}

We adopt the classification framework from OVMM~\cite{homerobotovmm} to categorize objects into three types: \emph{receptacle}, \emph{carriable}, and \emph{other}. \emph{Receptacles} are defined as flat, horizontal surfaces within a household, such as kitchen countertops, that serve as placement targets for rearrangement tasks. \emph{Carriable} objects are those that can be moved by robotic agents; these objects are often found misplaced and require appropriate relocation. The \emph{other} category encompasses objects that do not qualify as either \emph{receptacle} or \emph{carriable}. Although these objects cannot hold items or be moved by robots, they provide essential semantic context for other objects and are integral to the environment's overall understanding.

In our setup, a Fetch robot is employed to explore the environment and manipulate objects. We specifically designate objects that the Fetch robot can physically manipulate within the OVMM framework as \emph{carriable}. Items that are potentially movable but do not support manipulation in OVMM, such as paintings, mops, or plants, are classified as \emph{other}. However, by constructing a more detailed initial scene graph and refining the categorization of rearrangement types, our algorithm can be extended to accommodate additional objects, such as counters with drawers, enhancing its generalizability across different environments.
\vspace{-290pt}

\section{Scene Graph Construction}
\subsection{Initial Scene Graph Construction}

If we don't have this vanilla Scene Graph(SG), given an RGB-D sequence where each frame is annotated with its corresponding room, we can use this RGB-D sequence to construct a high-quality scene graph of this house and get all the necessary information. The pipeline can be described as the following algorithm~\ref{alg:SG_construction}.

Given RGB-D frames $\{ I^{c}, I^{d} \}$ with poses $x$, we use the depth camera's intrinsic parameters to convert each frame's depth image into the point cloud. By fusing the RGB point clouds from all RGB-D frames, we obtain the RGB point cloud $\mathcal{P}$ of the entire scene.
After that, we employ the OVIR-3D method\cite{lu2023ovir}, using the Detic model\cite{zhou2022detecting} to perform instance segmentation $M$ on each RGB image in the sequence, generating multiple instance masks and extracting CLIP\cite{radford2021learning} features $F$ for each instance region $R$. These instance masks and CLIP features are back-projected onto the corresponding 3D point clouds, forming 3D candidate regions.
We then merge similar 3D candidate regions by comparing their IOU and feature similarity. This process constructs 3D instances and determines the bounding box(bbox, OBB type) of each 3D instance in the point cloud. Ultimately, we obtain a set of 3D instances $\{{N}_1, ..., {N}_n\} \in \mathcal{N}$ of the point cloud.

\begin{algorithm}[t]
\caption{Process RGB-D frames to construct scene graph}
\label{alg:SG_construction}
\begin{algorithmic}[1]

\Statex \textbf{Input:} RGB-D frames $\{ I^{c}, I^{d} \}$ and corresponding pose $x$
\Statex \textbf{Output:} Hierarchical Scene Graph $\mathcal{S}$ with nodes $N$, edges $H$

\State $\mathcal{P} \leftarrow $ \textit{AlignWithPose}$(I^{d}, x)$;

\For{each RGB frame $I^{c}_{t}$ in $I^{c}$ }
    \State $M_{t} \leftarrow $ \textit{ImageSegmentation}$(I^{c}_{t})$;
    \State $F_{t} \leftarrow $ \textit{FeatureExtraction}$(M_{t})$;
    \State $R_{t} \leftarrow $ \textit{ProjectTo3D} $(F_{t},M_{t}, P) $;
\EndFor
\State $\mathcal{O} \leftarrow$ \textit{MergeCandidateToInstances} $(R)$;
\For{each 3D instance $O_{i}$ in $\mathcal{O}$}
    \State $N_{i} \leftarrow $ \textit{AssignWithOVMM}$(O_{i}, F_{i})$ ;\space \space \space // Assign the instance ID, category and rearrangement type 
    \For{each 3D instance ${O}_j (j \neq i) $ in $\mathcal{O}$ }
        \State $H_{i, j} \leftarrow $ \textit{GetRelationship} $(O_{i}, O_{j})$;
    \EndFor
\EndFor
\For{each 3D instance $O_{i}$ in $\mathcal{O}$}
    \State $\Omega{(N_{i})} \leftarrow $ \textit{SelectKeyFrame} $(M_{i},H_{i})$ ;
\EndFor
\end{algorithmic}
\end{algorithm}

We construct the Scene Graph based on these 3D instances $N$. Each 3D instance represents a node in the scene graph, containing the instance's ID as name. 
We use the CLIP feature $F$ of each 3D instance $N$ to find the most similar object label in the  ImageNet21K dataset\cite{deng2009imagenet}, assigning it as the category of the 3D instance and adding it to the node. After that, we calculate the number of visible pixels of each 3D instance in all RGB-D images, creating an object-to-image-pixel matrix $\textbf{P} \in \mathbb{R}^{N \times M}$ for room labeling and object keyframe $\Omega(N)$ selection. For each object, we assign the room $r$ annotation of the RGB frame with the highest number of visible object pixels as the room where the object is located and add this information to the node $N$. 
Furthermore, using the mapping function provided by OVMM\cite{homerobotovmm}, we determine each instance's rearrangement attribute (e.g. carriable, receptacle, or others) based on its category and include this attribute into the SG.

For the edge relationships $H$ in the scene graph, we calculate two types of spatial relationships: "near" and "on-supporting", using the bounding boxes (bbox) of these 3D instance pairs.
To find "on-supporting" relationship, we need to calculate the vertical spatial relation between pairs of objects. When the 2D IoU of two objects' bounding boxes projected on the XY plane is greater than 80\% while the 3D IoU is 0, we consider that one object is above the other one. If the relative height difference is less than 0.2 meters, we consider the lower object A to support the upper object B, and the upper object B to be on the lower object A. Additionally, since we are constructing the SG based on RGB-D images, and objects seen are generally not inside other objects, we exclude the "in" relationship. If one object C contains more than 80\% of the volume of another object D, we also consider object C to support object D, and object D to be on object C. For other cases, we use a predetermined spatial distance threshold to identify all pairs of 3D instances whose distance of closest vertices is less than this threshold and consider pairs of objects whose distance is less than this threshold to be in a "near" relationship. These relationships are assigned as directed edges in the SG, and also filled into a pair of nodes as attributes of spatial relationships with surroundings, completing the edge construction.

\subsection{Hierarchy Construction} 
With the given "room" attribute of each node in the SG, we can easily construct an object-room hierarchy for the scene graph.
The construction method for areas is as follows:
We use spectral clustering~\cite{li2007noise,von2007tutorial} to get areas in a room.
Based on the existing scene graph, we create a matrix $\textbf{G} \in \mathbb{R}^{n \times n}$ to represent the sub-graph in a room.
\begin{equation}
\textbf{G}_{i, j} = \begin{cases}
1 & distance(\textbf{O}_i,\textbf{O}_j)<\tau \\
0 & \text{otherwise}
\end{cases},
\label{eq:sg_edge}
\end{equation} 
where n is the number of instances in this room, $distance(\cdot)$ calculates the distance between two objects' bounding boxes, and the distance threshold $\tau$ is set to $2$ meters. Using this matrix, we calculate its normalized Laplacian matrix $\textbf{L}$ and then calculate the eigenvalues $\{\lambda_1, ..., \lambda_n\}$ of $\textbf{L}$. Follow~\cite{li2007noise,von2007tutorial}, we estimate the number of clusters $k$
by calculating the largest gap between the $\lambda_k$ and $\lambda_{k+1}$ by the following formula. 
\begin{equation}
k = {\arg\!\max}_{k} \{\mid\lambda_{k+1} -\lambda_k\mid \},
\end{equation}
 We then use spectral clustering to divide these objects into k groups. These k groups of objects are considered as k areas in the room. With these areas, we can construct the "object-area-room" hierarchy and convert our initial SG
 into a Hierarchical 3DSG, used for context aggregation.

We employ a method analogous to the one used for nodes to select a key frame for each area. However, unlike the approach for nodes, which focuses on the object itself and its immediate neighbors, this method involves identifying all objects within a specified area and calculating the sum of their projected pixel numbers. The frame that exhibits the maximum sum of projected pixels is then selected as the key frame for that area.


\subsection{Prompt design for affordance analysis}

The whole pipeline of scene graph enhancement with affordance analysis can be demonstrated as the following algorithm~\ref{alg:SG_affordance}:
\begin{algorithm}
\caption{Scene Graph Enhancement with Affordance Analysis}
\label{alg:SG_affordance}
\begin{algorithmic}[1] 
\Statex \textbf{Input:} Hierarchical Scene Graph $\mathcal{S}$ with nodes $N$, edges $H$ , areas $a$ and , rooms $r$
\Statex \textbf{Output:} Updated affordance ${A}^{updated}$

\For{each node $N_i$ in the $\mathcal{S}$}
    \State $A^{local}_{i} \leftarrow $ \textit{LocalAffordanceAnalyze}$(N_{i})$
\EndFor
\For{each room $r_i$ in the $\mathcal{S}$}
    \For{each area $a_k$ in the $r_{i}$}
        \State $\phi_{k} \leftarrow $ \textit{AreaAnalysis}$(a_{k}, r_{i})$
    \EndFor
    \State $\psi_{i} \leftarrow $ \textit{AggregateAreaAnalysis}$(\phi_{0:k})$

    \For{each node $N_j$ in the $r_{i}$}
        \State $H^{updated}_{j, *} \leftarrow$ \textit{UpdateSemanticEdges} $(A^{local}_{j}, \psi_{i})$
    
        \State ${A}^{updated}_{j} \leftarrow$ \textit{UpdateAffordance}$(A^{local}_{j}, \psi_{i}, H^{updated}_{j, *})$
    \EndFor 
\EndFor
\end{algorithmic}
\end{algorithm}

After obtaining the well-constructed hierarchical scene graph $\mathcal{S}$, we first traverse each node to perform a local affordance analysis $A^{local}$. For each receptacle and other object in the set $S$, we input all the textual contents related to its neighborhood and organize them into a descriptive text with the following template:

``\emph{The name of this receptacle is [], its category is [], it is located in the room [], its relationships with surrounding objects are: it is near these objects: [], it supports these objects: [], it is supported by these objects: [].} ''

Note that if the given object does not have a certain type of edge, then the corresponding description will not be generated.
Then we combine the template with its associated key frame and design a prompt for a large language model (LLM) to analyze the functionality of the object reflected in this scene. The LLM will follow a Chain of Thought (CoT) reasoning rule. First, it describes the object's geometry and position, then the relationships with surrounding objects, next it analyzes the object's functionality within the room, and finally, it assigns a better fine-grained category to the object.

Note that when conducting local context analysis, the information enhancement for the objects recognized as "carriable objects" is a bit different, as their local context within the environment frequently changes. Thus we simplify the analysis without considering the position and relationship.
For carriable objects, we input the category of the object as well as its key frame image, requiring LLM to output the affordance of the object with the following information: \emph{Geometry \& Functionality} and \emph{Fine-grained Category}. 

We facilitate the analysis of areas $\phi$ to capture global context $\psi$. For every area $a_k$, the \emph{categories} and \emph{fine-grained categories} of all present objects in the area, along with the selected key frame image $\omega(N)$ of the area, are inputted as prompt. The LLM generates outputs following a fixed template that includes  \emph{Description} and  \emph{Fine-grained Name}. This process is replicated across all areas within each room. The aggregated results from these analyses are then synthesized to establish the global context of the room for subsequent detailed analysis.

After obtaining the global affordance, for each object, we combine its local affordance $A^{local}$ and the global context of the room $\psi$ it is in as \emph{Additional information} to form a prompt input for LLM. The model is instructed to extract meaningful relationships using the format: \emph{Objects that have functional relationships} and \emph{additional functional edge}. We use the latter as the content for a semantic edge, constructing new semantic edges between the object and the objects that have functional relationships.

During the final phase of affordance updation, we take the global context of the room $psi$ in which the object is located, along with the newly constructed semantic edges $H^{updated}$, as \emph{Additional information}. This is combined with the local affordance and input into LLM to update the affordance. The updated affordance $A^{updated}$ will have the same format as the local affordance $A^{local}$. 
Note that we do not use the global context to update the affordance of carriables for the same reasons identified in the local analysis.

\section{LLM scorer prompting}

\subsection{misplacement detection}

To detect misplaced objects, we individually score all the carriables in the scene. For each carriable, we compile a scoring prompt from the description of the object, the context-induced affordance of the receptacle where it is located, and the task instruction "tidy the house." This prompt is used by LLM to assess the suitability of the object's placement on a scale from 0 to 100.
Recognizing that scoring objects individually might introduce variability due to the LLM's lack of consistent criteria, we incorporate a fixed standard within the prompt to guide the LLM's evaluations:

\begin{itemize}
    \item 
    100 points: the placement perfectly meets the task requirements;
    \item 
    0 points: the placement contradicts the task requirements and might negatively impact the task, and thus needs to be rearranged;
    \item 
    50 points: It is difficult to judge whether rearranging this object is related to the task.
\end{itemize}
This method is designed to standardize scoring while minimizing human bias regarding the evaluation criteria.

\subsection{placement planning}

We adapted our LLM-based misplacement detector with a slight modification in the prompt to facilitate Receptacle Candidate Retrieval. The input components remain consistent with the misplacement detector, including the task instruction "tidy the house," the description of the carriable object, and the receptacle’s context-induced affordance. However, we have tailored the scoring criteria in the prompt as follows: 
\begin{itemize}
    \item 
    100 points: The LLM identifies this receptacle as the most appropriate location in the entire house for placing the carriable to fulfill the task; 
    \item 
    0 points: The LLM deems this receptacle entirely unsuitable for placing the carriable, under any circumstances from the task’s perspective; 
    \item 
    50 points: The LLM considers this receptacle a plausible option only under specific conditions.
\end{itemize}
From the highest-scoring receptacles identified, we select the top-k candidates and analyze their context-induced affordance along with the query, which includes the description of the carriable and the task instruction, to generate placement decisions. The LLM then determines the optimal placement target, providing the outcome in the following format: 
\begin{itemize}
    \item 
    Best receptacle: the name of the most suitable receptacle among the candidates; 
    \item 
    Analysis: the rationale for its preference over other options.”
\end{itemize}

\section{Experiment Setting}

\subsection{Benchmark and evaluation metrics}
\paragraph{Public dataset.} We evaluate our method for the object rearrangement planning task on a benchmark dataset introduced by TidyBot~\cite{wu2023tidybot}. This benchmark is comprised of 96 scenarios, which are defined in 4 room
 types (living room, bedroom, kitchen, and pantry room). Each scenario contains 2–5 receptacles (potential places to put objects, such as shelves, cabinets, etc.) and 4 example object placements are given per receptacle. The success of this benchmark can be measured by the object placement accuracy.
\paragraph{Our context-oriented dataset.} Given that the existing TidyBot dataset only contains single-room scenes with simple layouts, the annotated ground truth of the object placement can not be based on the analysis of the scene context but more on the subjective bias of the user. To make a more fair and challenging evaluation, we contribute a novel benchmark of 8 multi-room scenes with complex scene layouts by annotating the Habitat Synthetic Scenes Dataset (HSSD 200)~\cite{khanna2023hssd}, which contains 36 rooms, 105 areas, and 188 receptacle objects. Among these, we selected 100 carriable objects over 60 categories and annotated their ground-truth placements with the help of seven skilled volunteers. 
They are required to control a robot, navigating each scene to identify the most appropriate receptacle for each new object, with a minimum of one and a maximum of five options. As all scene information was readily accessible, the provided annotations accurately reflect human commonsense about object placement. After gathering all annotations, we calculated the majority vote for each receptacle per object in each scene, generating a ranked list of receptacle choices.

\paragraph{Evaluation metrics and configurations.} When the household rearrangement task is complex, we measure the performance of our method based on misplaced object detection and rearrangement planning respectively. 
\begin{itemize}
    \item \textbf{Misplaced object detection:} We randomly place carriable objects into the benchmark scenarios to generate more than 4000 messy scenes for evaluation. When the ground-truth object placements are available, we measure the detection performance based on accuracy, recall, precision, and F1 score following most existing works~\cite{padilla2020survey}.
    \item \textbf{Rearrangement planning:} Given a set of carriable objects for a specific scene, we measure the rearrangement performance based on the corresponding 8 best placements for each carriable object given by different methods. The metric for the evaluation here is Normalized Discounted Cumulative Gain (NDCG)~\cite{jarvelin2002cumulated}.
\end{itemize}

\section{Ablation Study}
\subsection{Key Frame Selection}
\begin{figure}[t]
\centering{\includegraphics[width=0.5\textwidth]{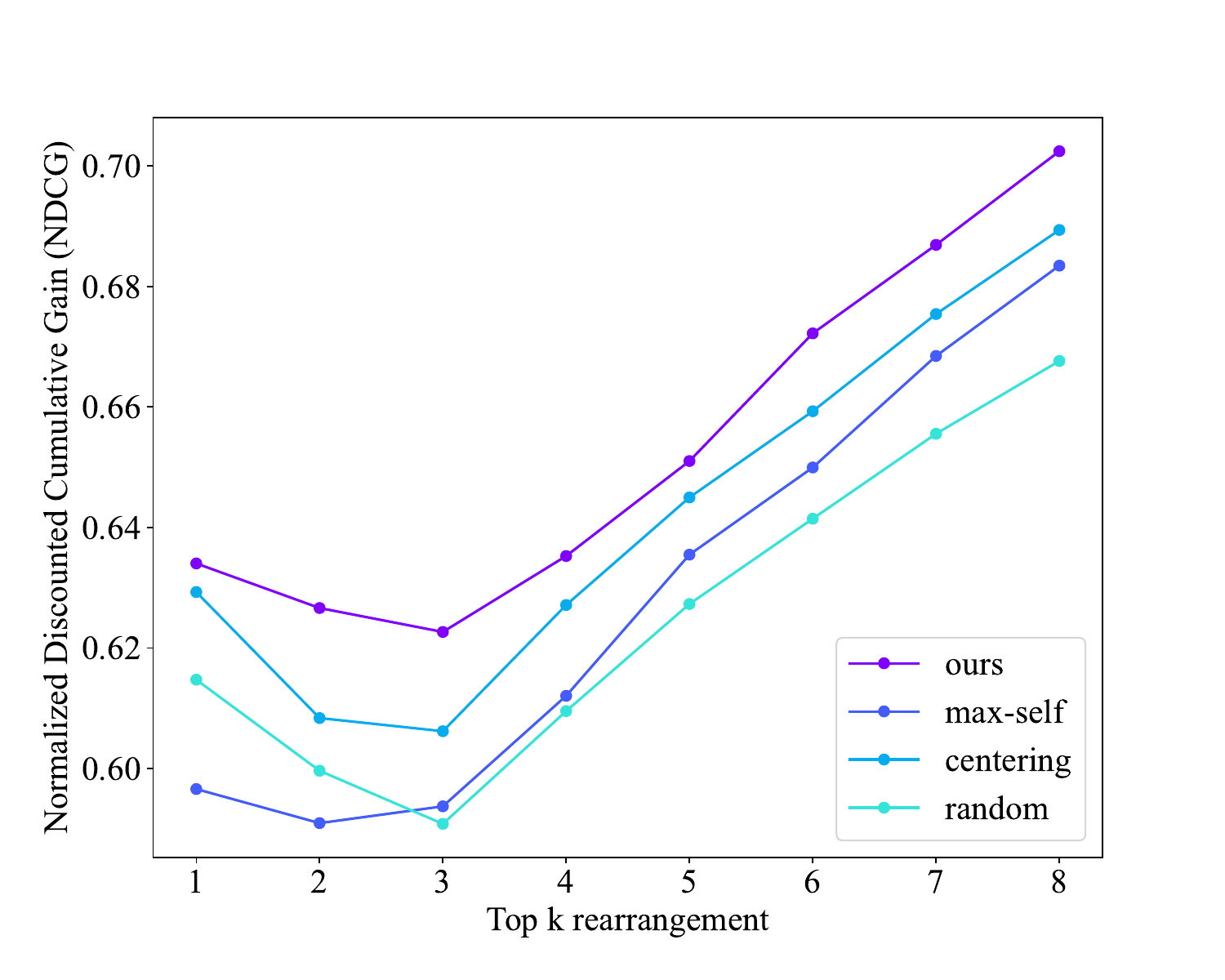}}
\caption{Ablation study on the selection of key frame via measuring the NDCG performance.
}
\label{fig:keyframe}   
\end{figure}

In this paper, we select a key frame for each object from an RGB sequence by calculating the total number of pixels occupied by the object and its neighbors in the image.
To evaluate the impact of key frame selection for each node in the scene graph on affordance analysis, we conducted an ablation study focusing on different key frame selection strategies. We compared three methods that only consider the number of pixels occupied by the object itself in the image against the method discussed in the paper. These methods are:
\begin{itemize}
    \item \textbf{Random selection:} Choose a frame randomly from all frames where the pixel count is not less than 100 pixels as the key frame.
    \item \textbf{Centering selection:} From all frames where the pixel count is not less than 100 pixels, select the frame where the object is closest to the center of the image as the key frame.
    \item \textbf{Max-self selection:} Select the frame where the object itself occupies the highest percentage of pixels in the image as the key frame.
\end{itemize}
We conducted affordance analysis based on these three different key frame selection strategies and performed experiments on misplacement detection and placement planning. The results, as shown in Figure ~\ref{fig:keyframe}, indicate that our method performs the best. While the centering-selection method also aims to preserve as much context information of the object by positioning it at the center of the image, it often results in the object being farthest from the camera when centered. This leads to the object appearing very small in the selected key frame, and the overall density of information about the object and its surrounding context in the image is also reduced. Conversely, the max-self method ensures that the object occupies a significant portion of the image. However, this approach often results in the object appearing at the corner or on one side of the image, consequently missing much of the surrounding information.

\subsection{Top K Retrieval}

\begin{figure}[t]
\centering{\includegraphics[width=0.5\textwidth]{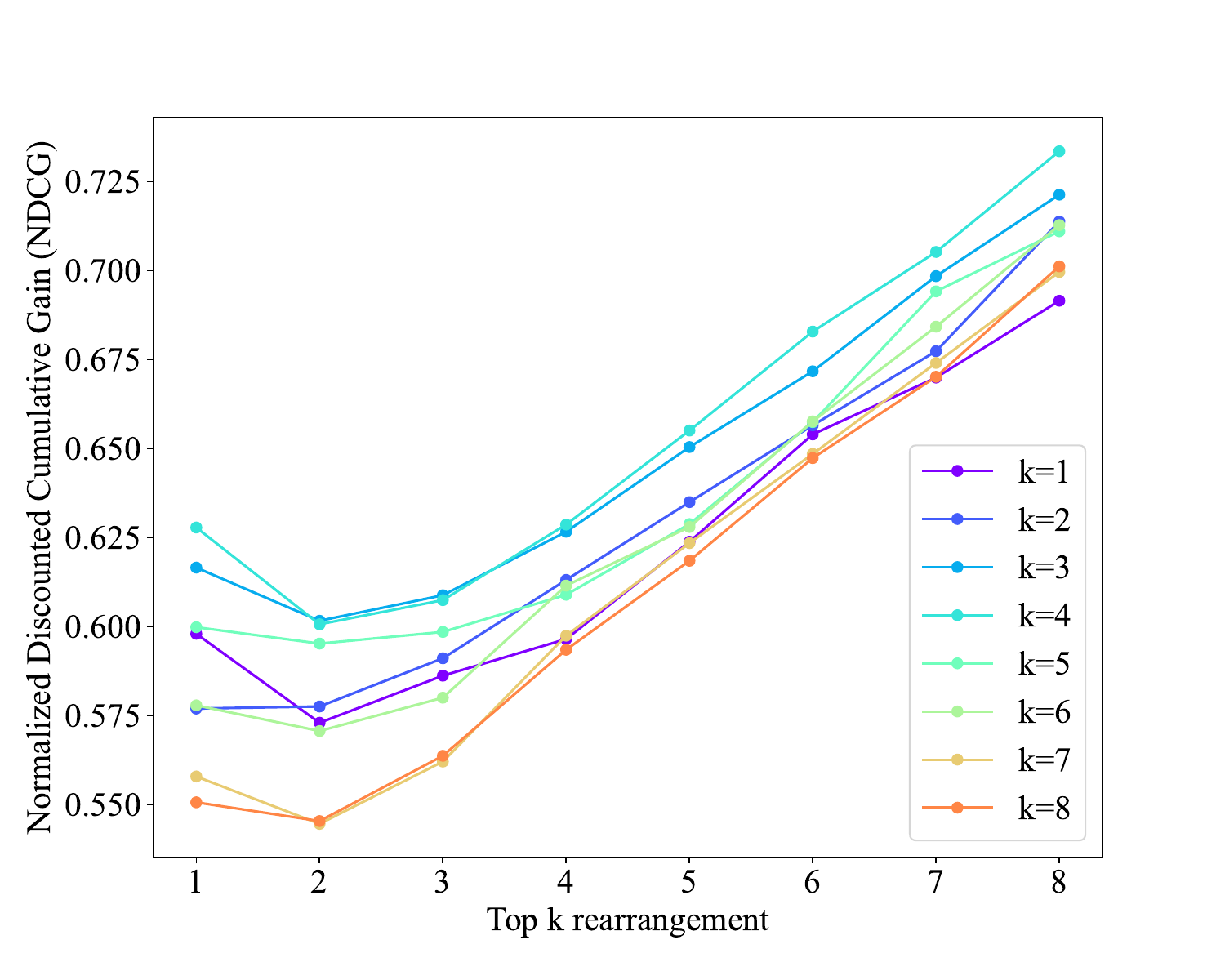}}
\caption{Ablation study on the number of retrievals $k$ via measuring the NDCG performance.
}
\label{fig:topk}   
\end{figure}

We also conducted an ablation study on the number of k in receptacle candidate retrieval within Object Rearrangement Planning to assess the impact of different k numbers on decision generation performance. We conducted experiments with k values ranging from 1 to 8 and carried out tests. The results are shown in Figure ~\ref{fig:topk}. results showed that performance initially increased and then decreased, peaking at k=3, 4. 
We consider that the decrease in decision-making performance when k is too large may be attributed to the increased token length as input and the progressively more challenging task of having to compare a greater number of objects against each other. This complexity might overwhelm the LLM's capacity to effectively process and prioritize among a larger set of options, thereby reducing the efficiency and accuracy of its decision-making.

\subsection{Score Prompting}

\begin{table}[t]
\caption{Ablation study on different scorer prompting aproaches via evaluating the misplacement detection performance on our context-oriented dataset. 
}
\label{tab:score_prompting}
\resizebox{0.9\linewidth}{!}{
    \begin{tabular}{l|llll}
\hline
Prompting Aproach      & accuracy & recall & precision & F1             \\ \hline
Ours                   & 0.829    & 0.880  & 0.908     & \textbf{0.894} \\
Fixed example          & 0.793    & 0.780  & 0.952     & 0.857          \\
Random example         & 0.820    & 0.826  & 0.944     & 0.882          \\
Self-generated example & 0.804    & 0.811  & 0.937     & 0.870          \\
No calibration         & 0.798    & 0.797  & 0.942     & 0.865          \\ \hline
\end{tabular}
    }
\end{table}

In our study, we tasked the LLM with individually scoring each placement, utilizing a straightforward, fixed standard within the prompt to calibrate the LLM’s evaluations. Given the robust contextual learning capabilities of large models, we hypothesized that incorporating a variety of scoring examples could enhance the model's accuracy. To assess the effectiveness of different prompting methods on the LLM’s scoring performance, we conducted a comparative analysis using the following approaches:
\begin{itemize}
    \item \textbf{Fixed example:} We provided two human-designed fixed scoring cases as examples to instruct the model.
     \item \textbf{Random example:} For each scoring instance, we randomly selected two examples from an existing library of scoring cases.
      \item \textbf{Self-generated example:} We used two scoring outcomes previously generated by the LLM itself as references.
       \item \textbf{No calibration:} We gave no example and omitted score calibration.
\end{itemize}
The experimental results, depicted in Table ~\ref{tab:score_prompting}, revealed that the simplest calibration method yielded the best outcomes. From the table, it is evident that the Fixed example, Random example, and Self-generated example methods all had lower accuracy and higher precision compared to our method. This demonstrates that the introduction of examples not only failed to enhance the robustness of the LLM's scoring but also introduced more human bias. Meanwhile, the No Calibration method, due to the lack of score calibration, obtained the lowest F1 score. This finding underscores the potential drawbacks of overly complex calibration processes and the importance of a balanced approach to maintaining model objectivity and effectiveness in scoring.

\section{applications}
\begin{figure*}[t]
\centering{\includegraphics[width=0.98\textwidth]{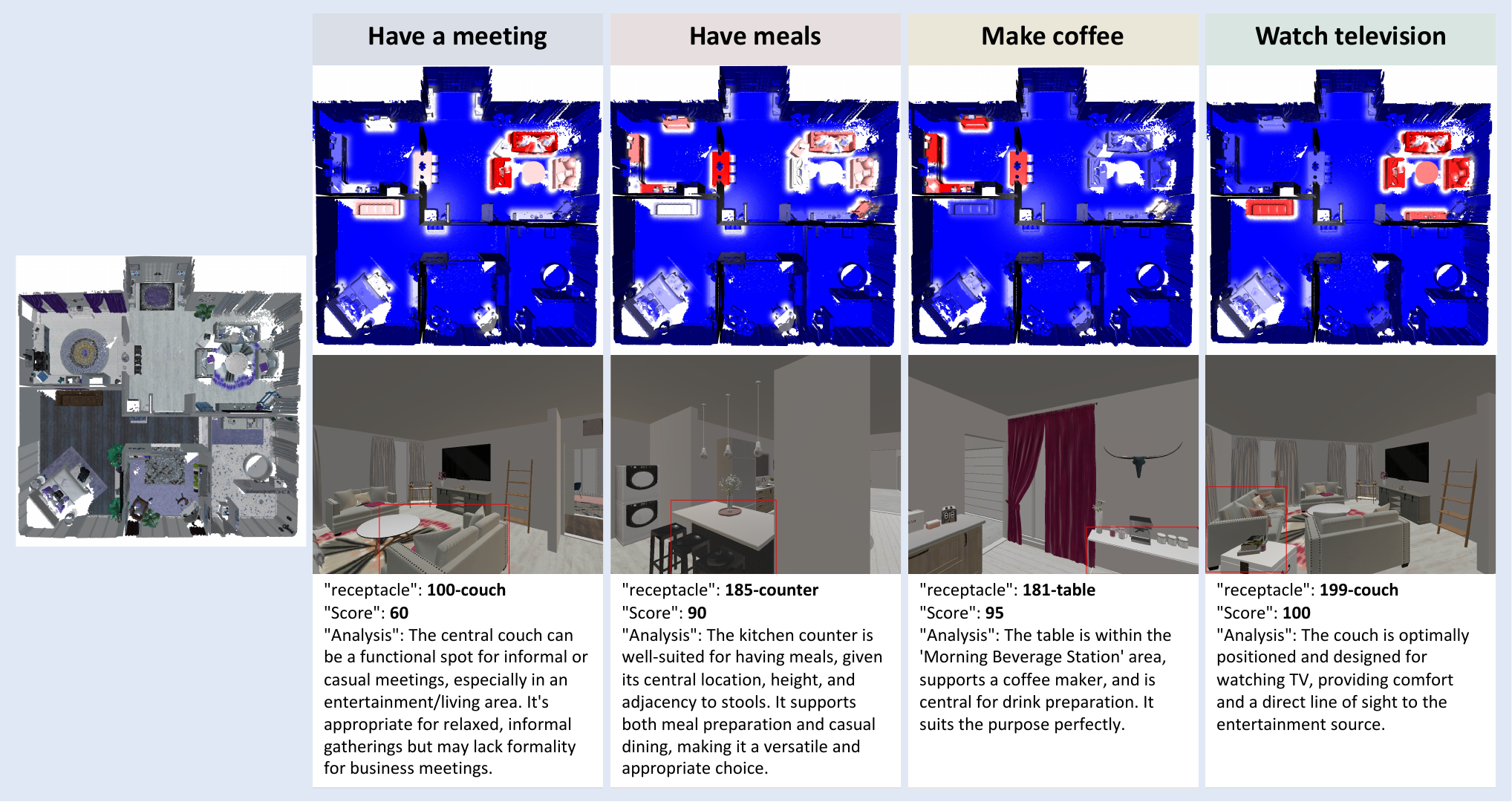}}
\caption{Examples for generating task-specific human-scene interaction heatmaps. By inputting a human action or activity instead of carriable affordance, our LLM can assess the relevance of an object's affordance to the specified human activity or action, and generate a heatmap based on this analysis. In the heatmap, red indicates a high relevance between the object and the activity, while blue indicates minimal relevance. This method allows for a visual representation of how different objects in a scene might interact with or support the human activity in question, providing intuitive insights for planning and interaction design.}
\label{fig:heatmap}
\end{figure*}

Our work involves high-level scene analysis dedicated to the in-depth mining of scene affordances based on both contextual analysis and common-sense reasoning with the help of large language models. The concept of affordance graph learning has potential applications in a variety of tasks beyond house tidying. Our affordance-enhanced graph (AEG) paired with an LLM scorer can accomplish diverse tasks such as scene editing, determining optimal positions for new objects, and making furniture recommendations through different inputs. In this section, we will demonstrate an example by generating a task-specific human-scene interaction heatmap by inputting a human activity or action. 

As shown in Figure ~\ref{fig:heatmap}, by changing the task instruction from "tidy the house" to "find a suitable place for a human activity," and replacing the input of carriable affordance with different human activities, the LLM can analyze and score the relevance between all objects in the house and the specified human activity. Using these scores, we can generate a task-specific human-scene interaction heatmap that illustrates the potential interactions between people and the scene. This human activity could be abstracted to "have a meeting" or be as specific as "make coffee." This process visually interprets how different elements in a scene may interact with human actions, thereby providing actionable insights for spatial arrangement and interaction design.



\section{LLM Prompt}
We show all input prompts passed to the LLM in Figure~\ref{fig:p1}, Figure~\ref{fig:p2}, Figure~\ref{fig:p3}, Figure~\ref{fig:p4}, Figure~\ref{fig:p5}, and Figure~\ref{fig:p6}.

\begin{figure*}[h]
    \centering
    \begin{tcolorbox}[colback=gray!10,
                  colframe=black,
                  arc=1mm, auto outer arc,
                  boxrule=1.0pt,
                 ]
    You are an intelligent home service agent that is professional in understanding contextual information of room layouts and analyzing objects usages in a particular house. \\
    Given a short semantic description as well as a image of a receptacle (the receptacle object is highlighted by a red bounding box). You need to: \\

    1. Describe the geometry and position of this receptacle in the house \\
    2. Describe the relationship between this receptacle and its surrounding objects, including objects it is supporting. (ignore the decorative objects) \\
    3. Based on your previous description, analyze the unique usage of this object as a receptacle and describe some possible items that can be placed on this receptacle. \\
    4. Assign this receptacle a new fine-grained Category as its unique characteristic in the house \\

    Output your analysis as following format: 
    
    \texttt{"""} \\
    1. ``Geometry \& Position": [your description about Geometry \& Position] \\
    2. ``Relationship": [your description about Relationships] \\
    3. ``Unique Usage": [your analysis for Unique Usage ] \\
    4. ``Fine-Grained Category": [the Fine-grained Name you give] \\
    \texttt{"""}
    
    Make your output concise. (It would be great if your output is under 150 words)
    \end{tcolorbox}
    \caption{Local context-induced affordance analysis}
    \label{fig:p1}
\end{figure*}

\begin{figure*}[h]
    \centering
    \begin{tcolorbox}[colback=gray!10,
                  colframe=black,
                  arc=1mm, auto outer arc,
                  boxrule=1.0pt,
                 ]
    You are an intelligent home service agent that is professional in understanding contextual information of room layouts and analyzing objects usages in a particular house. \\
    You will receive a text description of a list of objects that locate in an area of a particular room, se well as an image of this area. \\
    You need to analyze and summary the unique functionality of this area among areas in the particular room. \\

    Output your analysis as following format: \\
    \texttt{"""} \\
    1. ``Name": [assign a name of this given area in the room] \\
    2. ``Description": [Describe the layout and functionality of this area in one sentence] \\
    \texttt{"""}
    
    Make your output concise. (It would be great if your output is under 50 words)
    \end{tcolorbox}
    \caption{Area analysis}
    \label{fig:p2}
\end{figure*}
\begin{figure*}[h]
    \centering
    \begin{tcolorbox}[colback=gray!10,
                  colframe=black,
                  arc=1mm, auto outer arc,
                  boxrule=1.0pt,
                 ]
    You are an intelligent home service agent that is professional in understanding contextual information of room layouts and analyzing objects usages in a particular house. \\
    Given a short semantic description of a receptacle from the scene graph of a particular room as well as some additional information about this room. You need to: \\
    
    1.  analyze the description and find some objects from the additional information that have relationship with the given receptacle in functionality perspective \\
    2. add additional semantic edge between the finded objects and the given receptacle in the scene graph. \\

    Output your analysis as following format: \\
    \texttt{"""} \\
    1. ``Given Receptacle": [name of the given receptacle in scene graph] \\
    2. ``objects that have functional relationships": \\
        (1) [name of the first related object you find] \\
	   (2) [name of the second related object you find] \\
	   (3) ... \\
    3. ``additional functional edge": \\
	(1) [the functional relationship between receptacle and the first related object] (no more than 10 words) \\
	(2) [the functional relationship between receptacle and the second related object] (no more than 10 words) \\
	(3) ... \\
    \texttt{"""} \\
    Name of the related object should be consistent with the name in given object list from additional information.Ignore the decorative functional relationship. \\
    If you do not find any objects that have functional relationship, you can say ``(1) No object" in related subject. \\
    Make your output concise. (It would be great if your output is under 150 words)
    \end{tcolorbox}
    \caption{Semantic edge creation}
    \label{fig:p3}
\end{figure*}
\begin{figure*}[h]
    \centering
    \begin{tcolorbox}[colback=gray!10,
                  colframe=black,
                  arc=1mm, auto outer arc,
                  boxrule=1.0pt,
                 ]
    You are an intelligent home service agent that is professional in understanding contextual information of room layouts and analyzing objects usages in a particular house. \\
    Given a short semantic description as well as a image of a receptacle. You need to: \\
    
    1. Describe the geometry and position of this receptacle in the house \\
    2. Describe the relationship between this receptacle and its surrounding objects, including objects it is supporting. (ignore the decorative objects) \\
    3. Analyze the unique usage of this receptacle as a receptacle based on your previous contextual description and describe some possible interaction between human and this receptacle in the house. \\
    4. Assign this receptacle a new fine-grained Category as its unique characteristic in the house \\

    Output your analysis as following format: \\
    \texttt{"""} \\
    1. ``Geometry \& Position": [your analysis about Geometry \& Position] \\
    2. ``Relationship": [your analysis for Relationships] \\
    3. ``Unique Usage": [your analysis for Unique Usage ] \\
    4. ``Fine-Grained Category": [the Fine-grained Name you give] \\
    \texttt{"""} \\
    Make your output concise. (It would be great if your output is under 150 words)
    \end{tcolorbox}
    \caption{Context-induced affordance updation}
    \label{fig:p4}
\end{figure*}

\begin{figure*}[h]
    \centering
    \begin{tcolorbox}[colback=gray!10,
                  colframe=black,
                  arc=1mm, auto outer arc,
                  boxrule=1.0pt,
                 ]
        You are an intelligent home service robot tasked with house tidying.  you are professional in understanding the usage of the object in a particular house based on semantic and visual description and find a appropriate receptacle to place it for house tidying purpose. \\
        Given the semantic description of a carriable object to place and a possible receptacles in the house, you need to rate this receptacles on whether it is suitable to place the object for housekeep purpose. \\
        you can rate each receptacle at 0 to 100 score, where 100 means you think this receptacle might be the best position in the house to place the carriable object for house keep purpose , while 0 means that you do not recommend to place the object here for any reason. (As a reference, you can score the receptacle as 50 if you think it is only reasonable under some special conditions to place the object on this receptacle for kousekeep purpose.) Sometimes you will be given more reference to support your rating, these reference are placements that you have rated before .Make sure your scoring criteria are consistent compared to these references. \\
    
        Output your analysis as following format: \\
        \texttt{"""} \\
        1. ``name of the carriable": [name of the carriable object you need to place] \\
        2. ``name of the receptacle": [name of the give receptacle as the placement target] \\
        3. ``Score": [the score you give for this placement] \\
        4. ``Analysis": [your Analysis about why you give this placemnt this score for housekeep purpose] \\
        \texttt{"""} \\
        Make your output concise. (It would be great if your output is under 50 words)
    \end{tcolorbox}
    \caption{Task-relevant placement scorer}
    \label{fig:p5}
\end{figure*}

\begin{figure*}[h]
    \centering
    \begin{tcolorbox}[colback=gray!10,
                  colframe=black,
                  arc=1mm, auto outer arc,
                  boxrule=1.0pt,
                 ]
    You are an intelligent home service robot tasked with house tidying.  you are professional in understanding the usage of the object in a particular house based on semantic and visual description and find a appropriate receptacle to place it for house tidying purpose. \\
    Given the semantic description of carriable object to place and some possible receptacles in the house as placement target, you need to choose the best (the most suitable) place to place the carriable object on top of that place for house tidying purpose.Think this question in multiple perspective like feasibility, necessity and significance of the placement for house tidying purpose. \\
    Output your analysis as following format: \\
    \texttt{"""} \\
    1. ``The best receptacle": [the exact name of the best place among the given receptacles]. \\
    2. ``Analysis": [your Analysis about why it is the best one to place the object (why is it better than others)] \\
    \texttt{"""} \\
    Make your output concise. (It would be great if your output is under 100 words)
    \end{tcolorbox}
    \caption{Retrieval-augmented placement generator}
    \label{fig:p6}
\end{figure*}

\end{document}